# DC-SPP-YOLO: Dense Connection and Spatial Pyramid Pooling Based YOLO for Object Detection

Zhanchao Huang, Jianlin Wang*, Xuesong Fu, Tao Yu, Yongqi Guo, Rutong Wang
College of Information Science and Technology, Beijing University of Chemical Technology,
Beijing 100029, China

**Abstract:** Although the YOLOv2 method is extremely fast on object detection, its detection accuracy is restricted due to the low performance of its backbone network and the underutilization of multi-scale region features. Therefore, a dense connection (DC) and spatial pyramid pooling (SPP) based YOLO (DC-SPP-YOLO) method for ameliorating the object detection accuracy of YOLOv2 is proposed in this paper. Specifically, the dense connection of convolution layers is employed in the backbone network of YOLOv2 to strengthen the feature extraction and alleviate the vanishing-gradient problem. Moreover, an improved spatial pyramid pooling is introduced to pool and concatenate the multi-scale region features, so that the network can learn the object features more comprehensively. The DC-SPP-YOLO model is established and trained based on a new loss function composed of MSE (mean square error) loss and cross-entropy loss. The experimental results indicated that the mAP (mean Average Precision) of DC-SPP-YOLO is higher than that of YOLOv2 on the PASCAL VOC datasets and the UA-DETRAC datasets. The effectiveness of DC-SPP-YOLO method proposed is demonstrated.

## 1.Introduction

The object detection methods based on computer vision have been widely applied in the fields of security monitoring, autonomous driving, and medical diagnosis. The early object detection methods almost rely on key points, edges, or templates, of which the detection accuracy is low and the application range is limited. In this regard, feature extraction methods with better object expression such as Haar-like features, HOG (Histogram of Oriented Gradient) and LBP (Local Binary Patterns) were proposed and applied for object detection together with machine learning [1].

In 2007, Felzenszwalb et. at.[2] introduced the DPM (Deformable Parts Models) creatively, which got higher detection accuracy by a new object detection pipeline based on handcrafted futures and machine learning. After that, various object detection methods based on DPM were proposed and performed well in successive PASCAL VOC object detection challenges [3,4]. However, most of these methods needed to scan through the entire image for detecting the object regions by a sliding window, which resulted in the inefficient detection. Also, the further improvement of detection accuracy was restricted by the expression performance of handcrafted futures.

With the development of graphics processors and the enrichment of data resources, the convolutional neural network (CNN) with powerful image understanding and expression performance was proposed [5]. It was proved that object images could be classified more accurately by using the CNN features than using the handcrafted futures [6], which provided new ideas for object detection. In 2014, the R-CNN proposed by Girshick et. al. [7] employed CNN to extract rich features in object detection task for the first time, which got the state-of-the-art performance at that time.

At present, deep learning has become a research hotspot in the object detection field, and two types of deep learning object detection methods (the proposal-based methods and the regression-based methods) have been developed [8].

The proposal-based methods are developed from the R-CNN method. Aiming at the problem of inefficient detection caused by the object search strategy using sliding windows, the Fast R-CNN method [9] and the Faster R-CNN method [10] respectively utilized the selective search strategy and the region proposal network (RPN) to simplify the generation process of region proposal generation. Dai et. al. [11] proposed an R-FCN (Region-based Fully Convolutional Networks) to solve the problem that the ROI-wise subnetwork of Faster R-CNN did not share calculations in different region proposals. In the past two years, based on the Faster R-CNN method and the R-FCN method, RRPN (Rotation Region Proposal Networks) [12], R-FCN-3000 [13] and other proposal-based methods [14,15] have been proposed successively, and the object detection accuracy has been further improved. However, these methods all perform the region-proposal generation stage and the subsequent feature resampling stage respectively, which makes it difficult to meet the real-time requirements for object detection.

The first regression-based method, YOLO (You Only Look Once), proposed by Redmon et al. [16] in 2016 simultaneously predicted the coordinates of bounding boxes and classified the objects in an end-to-end neural network. Even if YOLO opened the door for the real-time object detection, it was still difficult to detect small-sized objects, and the error of bounding box coordinates was large as well. In this regard, Liu et. al. [17] proposed an SSD (Single Shot Multi-Box Detector) that introduced reference boxes and detected the object on multi-scale feature maps to improve the detection accuracy. Redmon and Farhadi [18] proposed the YOLOv2 method with higher accuracy and faster speed compared with the YOLO method. However, this method still used the Darknet19 backbone network with low performance of feature extraction and did not fully utilize the multi-scale region features, which constrained the further improvement of detection accuracy. Subsequently, the deep residual network (ResNet) was employed as the backbone network in DSSD (Deconvolutional Single Shot Detector) [19] and YOLOv3 [20] to get state-of-the-art detection accuracy. But on the other hand, the detection speed of these methods is severely degraded due to the more complex network. Zhou et. al. [21] proposed an STDN (Scale-Transferrable Detection Network) method. It introduced the DenseNet-169 [22] as the backbone network of SSD and got the detection accuracy close to that of DSSD method with the faster detection speed. Although many improved SSD methods have been proposed [23-26], but to our knowledge, the existing research on the improvements of YOLO series methods are still less.

Therefore, a DC-SPP-YOLO object detection method is proposed in this paper for improving the detection accuracy of YOLOv2 while keeping the real-time detection speed. The main contributions of our work are as follows. (1) The connection structure of the backbone network is optimized by the dense connection strategy for ameliorating the detection accuracy by strengthening the feature propagation and ensuring the maximum information flow in the network. (2) An improved spatial pyramid pooling structure is introduced to pool and concatenate the regional features on different scales in the same convolutional layer for the less location error when detecting the small objects. (3) A new loss function, consisting of the MSE loss for location and the cross-entropy loss for classification, is employed for the faster model training speed and the higher detection accuracy.

This paper is organized as the following. Section 2 reviews the related works on YOLOv2 method, backbone networks, multi-scale detection. Section 3 explains the DC-SPP-YOLO in detail. Section 4 gives the object detection process using DC-SPP-YOLO. Section 5 presents a series of experiments and discusses the results. Section 6 draws up the conclusions and makes prospect about the future work.

## 2. Related Works

### 2.1. YOLOv2 Method

The YOLOv2 object detection method [18] divides the input image into $S$×$S$ grids. Each grid predicts $K$ bounding boxes. The class-specific confidence of each bounding box is

$$\Pr(\text{Class}_i \mid \text{Object}) * \Pr(\text{Object}) * \text{IoU}_{\text{pred}}^{\text{truth}} = \Pr(\text{Class}_i) * \text{IoU}_{\text{pred}}^{\text{truth}} \tag{1}$$

Where $\Pr(\text{Object}) * \text{IoU}_{\text{pred}}^{\text{truth}}$ is the confidence that the bounding box contains objects; $\text{IoU}_{\text{pred}}^{\text{truth}}$ is the Intersection-over-Union between the predictions and the ground truth; $\Pr(\text{Class}_i \mid \text{Object})$ is the conditional probabilities that the object belong to $C$ classes [16]. Therefore, the predictions of YOLOv2 are encoded as an $S\times S\times(K\times(5+C))$ tensor.

The backbone network of the YOLOv2 extracts the object features by the down-sampling convolutional structure that is similar to the VGG network. When convolutional neural network forward propagates, the relationship between the $l$th layer and the $l$-1th layer is a function as following

$$\boldsymbol{x}^l = f\left(\boldsymbol{y}^l\right) = f\left(\boldsymbol{x}^{l\text{-}1} * \boldsymbol{w}^l + \boldsymbol{b}^l\right) \tag{2}$$

The input of the $l$th layer in the convolutional neural network is represented as $\boldsymbol{x}^l$. The activation function is $f(.)$. The intermediate variable is represented as $\boldsymbol{y}^l = \boldsymbol{x}^{l\text{-}1} * \boldsymbol{w}^l + \boldsymbol{b}^l$; where $\boldsymbol{w}^l$ is the weight of the convolution kernel, $\boldsymbol{b}^l$ is the bias parameter, * represents convolution.

When convolutional neural network backpropagates, the gradient of the loss function is

$$\boldsymbol{\delta}^{l-1}=\frac{\partial \boldsymbol{L}}{\partial \boldsymbol{y}^{l\text{-}1}}=\frac{\partial \boldsymbol{L}}{\partial \boldsymbol{y}^{l}}\cdot\frac{\partial \boldsymbol{y}^{l}}{\partial \boldsymbol{y}^{l\text{-}1}}=\boldsymbol{\delta}^{l} * \mathrm{rot180}\left(\boldsymbol{w}^{l}\right) \odot f'(\boldsymbol{x}^{l\text{-}2} * \boldsymbol{w}^{l\text{-}1}+\boldsymbol{b}^{l\text{-}1}) \tag{3}$$

In Eq. (3), $\mathrm{rot180}(.)$ represents the 180° counterclockwise rotation of the weight parameter matrix; $\odot$ is the Hadamard product; $\boldsymbol{L}(.)$ is the loss function. As the gradient propagates layer by layer in the network, the gradient represented by the product of the derivative of the activation functions and the weight parameters will become smaller and smaller. For example, the derivative of the Sigmoid activation function is $\left|f'(\boldsymbol{y}^{l\text{-}1})_{\mathrm{Sigmoid}}\right| \leq 1/4$, the initialized weights are usually less than 1; the gradient will vanish when it backpropagates in the network. Finally, the vanishing-gradient problem appears and results in low detection accuracy.

Besides, for multi-scale detection, the “Fine-Grained Features” strategy employed in YOLOv2 does not fully utilize the multi-scale local region features, which restricts the improvement of detection accuracy.

## 2.2. Backbone network

As the feature extractor, the backbone network plays a significant role in object detection. The accuracy and speed of object detection is directly related to the performance of the backbone network. The existing network design ideas mainly include “Repeat”, "Skip-connection" and "Multi-path".

Using the “Repeat” idea, the VGG network [27] employed a stack of small convolutional kernels instead of a single large convolutional kernel to deepen the network. It has been adopted as the backbone network by many popular object detection methods such as Faster R-CNN and SSD. Nevertheless, as the number of convolutional layers increased, further improvement of detection accuracy was restricted because of the vanishing-gradient resulted from the connection between convolutional layers within the VGG network.

Highway Network [28] and ResNet [14] was introduced the "Skip-connection" idea to deepen the convolutional neural network further while alleviating the vanishing-gradient problem. Subsequently, as the backbone network, ResNet was adopted by Faster R-CN, R-FCN, DSSD, and other methods. Compared with the methods using VGG network as the backbone network, these methods with better feature extraction improved the detection accuracy significantly, while the detection speed was severely degraded because of the extreme deep network. In 2017, Huang et. al. [22] presented the DenseNet with the dense connection structure of convolutional layers, which was faster and more accurate than ResNet on the image recognition task while alleviating the vanishing-gradient problem further. In STDN, DenseNet-169 was used as the backbone network. The object detection speed was significantly improved compared with the DSSD using the ResNet-101 as the backbone network [21].

The "Multi-path" employed by the Inception series network [29-32] and the Xception network [33] was also one of the main ideas for backbone network design and improvement. Li et. al. [34] adopted the improved Xception network as the backbone network in the proposed

Light-head R-CNN method and got better performance than YOLO and SSD on the COCO datasets.

Besides, convolutional neural networks such as MobileNet [35], SqueezeNet [36], ShuffleNet [37] compressed the network for higher speed. Even if the object detection methods using compressed networks were less accurate than the methods using larger backbone networks like ResNet and Inception, the detection speed was significantly increased and was applied widely in object detection tasks on mobile terminals.

At present, the replacement of advanced backbone networks based on "Skip-connection" or "Multi-path" to the VGG networks has become one of the main improvements on object detection tasks. The accuracy is ameliorated by strengthening the feature extraction and reusing the object features, but the detection speed also decreases. The DenseNet has the advantages of alleviating the vanishing-gradient and reusing the object features so that the STDN using the DenseNet can improve the detection accuracy with maintaining a fast detection speed. However, the detection speed of STDN is still lower than that of YOLOv2. Therefore, the dense connection will improve the detection accuracy and speed of YOLOv2 effectively.

## 2.3. Multi-scale Detection

Multi-scale detection is one of the significant research fields on CNN-based object detection. In recent years, a variety of multi-scale object detection methods have been developed, which are mainly divided into two categories: independent detection on multiple feature maps extracted by different layers of the networks, and fusing multiple feature maps extracted by different layers of the networks.

The method of independent detection on multiple feature maps was first adopted in the SSD proposed by Liu et. al. [17]. It was demonstrated better for detecting small objects than detecting objects on the feature map extracted by coarser top layers of the network. In 2016, Cai et. al. [38] improved the Faster R-CNN and detected objects on multi-scale feature maps, for which the receptive fields could adapt to multi-scale objects. This method provided a good performance in the scene where the scales of the object changed considerably. Yang et. al. [39] proposed the SDP (Scale Dependent Pooling) method, which pooled features from different convolutional feature maps according to the size of each proposal. In 2018, Li et. al. [40] used the scale-aware mechanism to weight and combine the prediction results of large and small size Sub-network according to the size of input proposal, which got the state-of-the-art performance on pedestrian detection.

The method of fusing multiple feature map improves the accuracy of multi-scale detection by fusing information from different scale feature maps and receptive fields. In 2014, the SPP (Spatial Pyramid Pooling Network) method presented by He K et. al. [41] pooled arbitrary size feature maps into fixed-size feature vectors, for which the CNN model didn't need to fix the size of the input images but also became robust for detecting multi-scale objects by fusing the multi-scale features. In 2017, different methods of fusing the multi-scale feature maps was reported to improve SSD and got better performances than SSD [19,23,42], for which the finer layers of the networks could utilize the contextual information learned from the coarser layers of the networks. Lin et. al. [43] took one step ahead and proposed the FPN (Feature Pyramid Network) method, in which a top-down lateral connection structure was designed based on the multi-scale pyramid structure inherent in deep convolutional neural networks, for increasing the accuracy of multi-scale detection. In 2019, Zeng et. al. [44] proposed the PPN (proposal pyramid networks) that avoids the traditional image pyramid structure and reduces the major computational complexity for fast face detection.

The multi-scale detection methods above detect objects independently on different feature maps or multi-scale feature maps fused by utilizing the global features from different convolutional layers of the networks to improve the detection accuracy. However, these methods do not make full use of the local region features on different scales from the same convolutional layer, and it is still difficult to detect small objects with rich local region features accurately.

# 3. Dense Connection and Spatial Pyramid Pooling Based YOLO

Based on the network of YOLOv2, we obtain the network of DC-SPP-YOLO by replacing a part of laminated convolutional layers with the dense connected convolutional layers and introducing the new space pyramid pooling block. The following is a further explanation of our improved dense connection and the improved spatial pyramid pooling in the Section 3.1 and Section 3.2 respectively. In the Section 3.3, the construction of DC-SPP-YOLO network and loss function are explained in details.

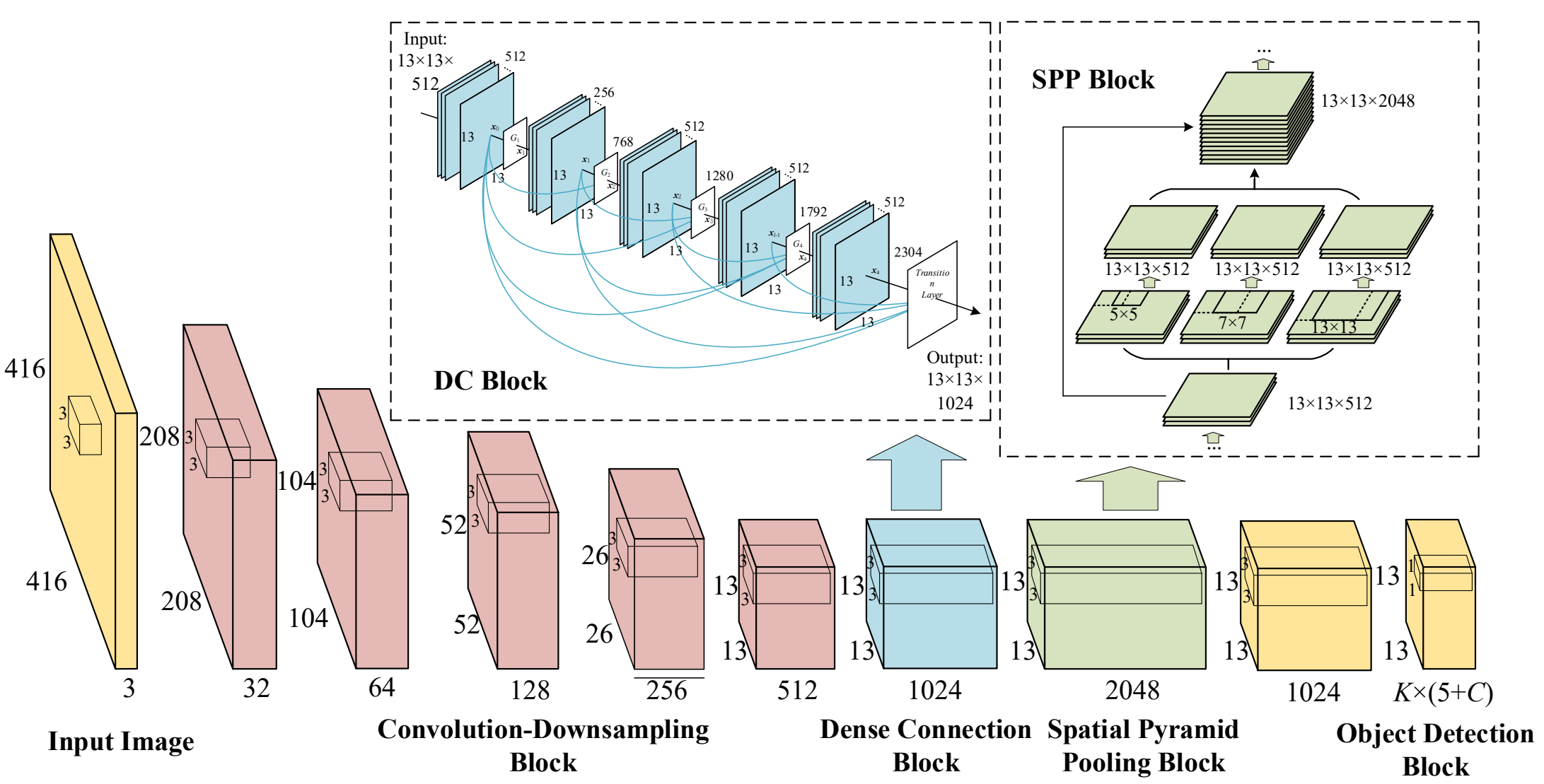

**Fig. 1.** The DC-SPP-YOLO Model.

## 3.1. Improved Dense Connection in YOLOv2

Considering the low ability of the backbone network on feature extraction and the vanishing-gradient in backpropagation, a dense connection structure of convolutional layers was employed to improve the accuracy of YOLOv2 by strengthening the feature extraction ability while ensuring the maximum information flow in the network.

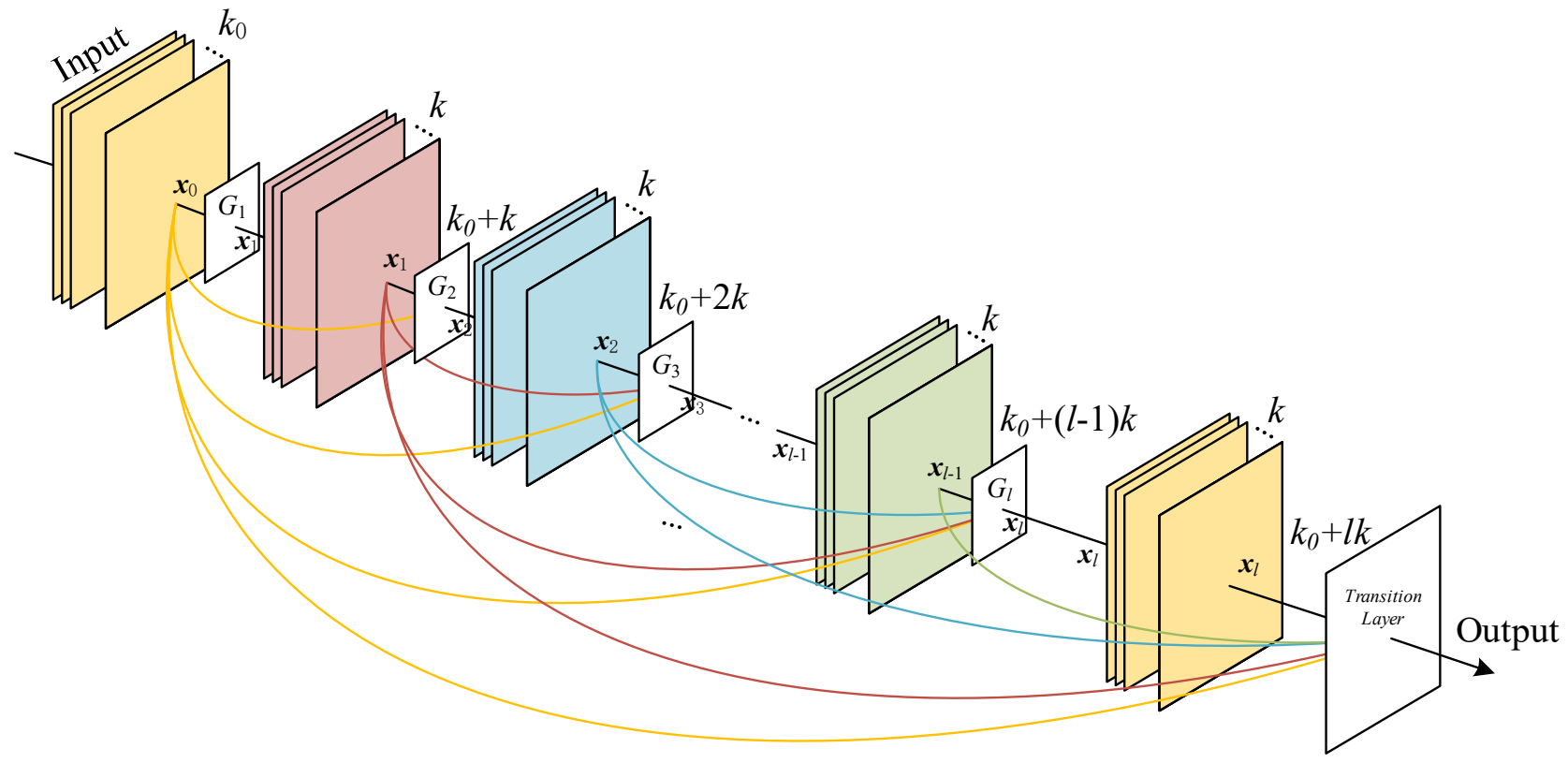

**Fig. 2.** The Dense Connection of the Convolutional Layers.

The dense connection structure of DC-SPP-YOLO in which the feature maps of the first $l$-1 layers are concatenated together and utilized as the input of the $l$th layer is shown in Fig. 2. When the network with dense connection structure forward propagates, the relationship between the $l$th layer and the $l$-1th layer is represented as

$$\boldsymbol{x}^{l} = f\left(\boldsymbol{y}^{l}\right) = f\left(\left[\boldsymbol{x}^{0}, \boldsymbol{x}^{1}, \cdots, \boldsymbol{x}^{l-1}\right] * \boldsymbol{w}^{l} + \boldsymbol{b}^{l}\right) \tag{4}$$

When convolutional neural network backpropagates, the gradient of the loss function is

$$\boldsymbol{\delta}^{l-1} = \boldsymbol{\delta}^{l} * \mathrm{rot}180\left(\boldsymbol{w}^{l}\right) \odot f'(\left[\boldsymbol{x}^{0}, \boldsymbol{x}^{1}, \cdots, \boldsymbol{x}^{l-2}\right] * \boldsymbol{w}^{l-1} + \boldsymbol{b}^{l-1}) \tag{5}$$

Compared with the derivative term $f'(\boldsymbol{x}^{l-2} * \boldsymbol{w}^{l-1} + \boldsymbol{b}^{l-1})$ of the activation function in Eq. (3), the derivative term $f'(\left[\boldsymbol{x}^{0}, \boldsymbol{x}^{1}, \cdots, \boldsymbol{x}^{l-2}\right] * \boldsymbol{w}^{l-1} + \boldsymbol{b}^{l-1})$ of the activation function in equation (5) always contains the input $\boldsymbol{x}^0$ and the output feature maps of the pr evious layers. Therefore, each layer of the CNN can obtain the input features and the gradient can be calculated directly from the loss function. It could alleviate the vanishing-gradient and increases the detection accuracy through the improvement feature propagation in the network.

Each convolutional layer of the DC (Dense Connection) block in DC-YOLO (Dense Connection Based YOLO) outputs $k$ concatenated feature maps. The $l$th layer of DC block outputs $k_0$+ $k$×($l$-1) concatenated feature maps, where the number of the input feature maps $\boldsymbol{x}^0$ is $k_0$.

Considering that detection speed was severely degraded due to the excessive number of residual connection layers in YOLOv3, only the last convolutional block which extracts the richer semantic features in the backbone network of YOLOv2 is improved to be dense connection block.

As shown in Fig. 3, the DC block has four dense connection units, which was composed of a 3×3 a 1×1 convolutional layer, and the increments of feature maps are set to be 256, 512, 512, and 512 respectively.

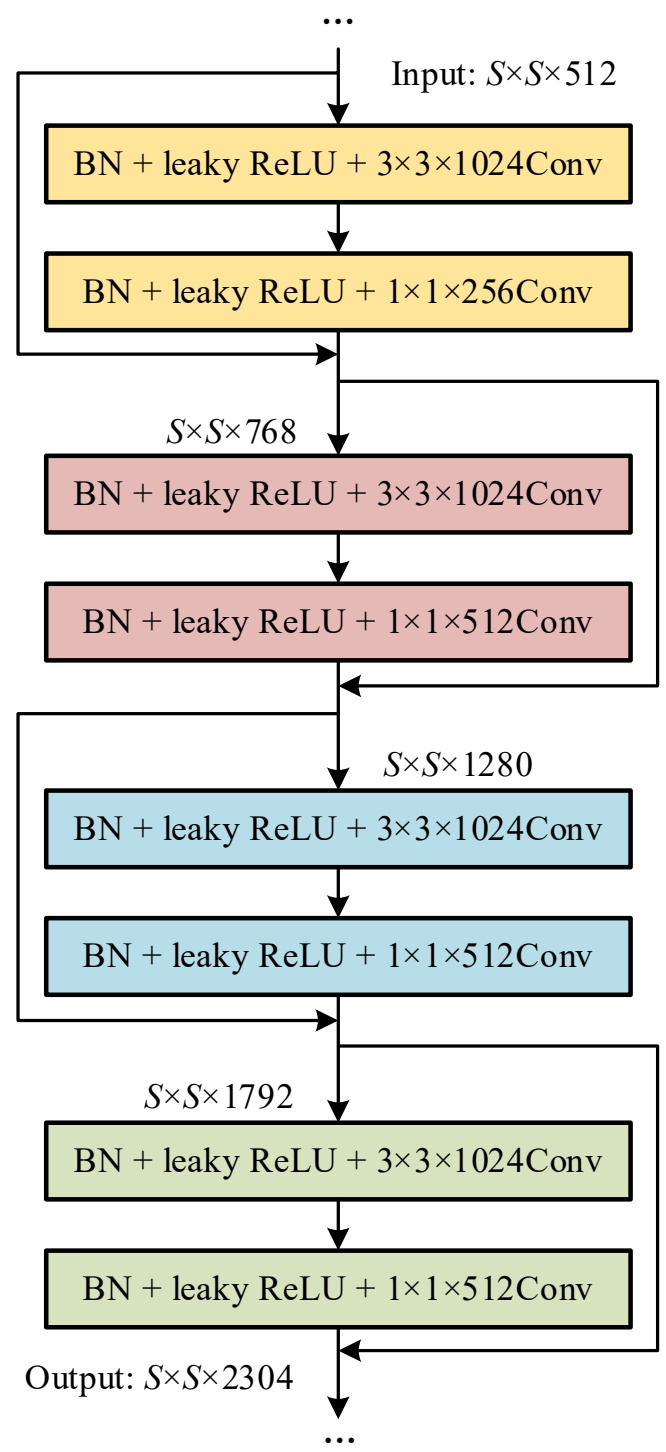

**Fig. 3.** The DC Block in DC-YOLO.

BN (Batch Normalization) [30] is added to solve the "internal covariate shift" and alleviate the vanishing-gradient problem and speed up the model training. The leaky ReLU activation function

$$y_i = \begin{cases} x_i & \text{if } x_i \geq 0 \\ \dfrac{x_i}{a_i} & \text{if } x_i < 0 \end{cases}, \quad a_i \in (1, +\infty) \tag{6}$$

is utilized for the nonlinearization of convolution. When the input is greater than 0 ($f'(y^{l-1})_{\text{leaky ReLU}} = 1$), the vanishing-gradient can be alleviated.While the input is less than 0 ($0 < f'(y^{l-1})_{\text{leaky ReLU}} < 1$),the dead neuron can be reduced compared to the ReLU activation function.

Considering that the DC block is in the deeper layer of the network, where the features extracted are fine-grained and the receptive field of each feature is also larger, using the larger convolution can extract the richer semantic features to describe the object better. Therefore, unlike the dense units with the "Bottleneck Layers" structure in DenseNet, each dense unit of DC-YOLO first extracts the object features by a 3×3 convolution to ensure more abundant fine-grained features are utilized. And then a 1×1 convolution is employed to reduce the number of input feature-maps. Nevertheless, this design of connection also leads to an increase in the number of model parameters.

Therefore, the nonlinear mapping function of each dense unit can be represented as BN-leaky ReLU-Conv(3×3)-BN-leaky ReLU-Conv(1×1). The DC block with eight convolutional layers replaces the original laminated convolutional block with four convolutional layers. Although the number of network layers was increased in a small amount, but the detection accuracy was improved while maintaining a fast detection speed.

**Table 1**

The comparison of DC-YOLO and YOLOv2.

| Method | BFLOP/s | mAP (%) on VOC 2007 | Speed (fps) |
|---|---|---|---|
| YOLOv2 | 29.371 | 76.8 | 67 |
| DC-YOLO | 39.383 | 77.6 | 58.9 |

In Table 1, BFLOP/s (Billion Floating Point Operations per Second) is used as an evaluation index to compare the model complexity of YOLOv2 and DC-YOLO, as well as the detection accuracy and speed of YOLOv2 and DC-YOLO on the PASCAL VOC 2007 dataset (see Section 5 for specific experimental settings). As shown in Table 1, the object detection accuracy of DC-YOLO is 77.6% on the PASCAL VOC 2007 dataset, which is 0.8% higher than that of the YOLOv2. Although the model complexity of DC-YOLO is increased by 10.012 BFLOP/s compared to YOLOv2, the detection speed only reduces by about 8.1 fps.This means that DC-YOLO still maintains a fast detection speed.

## 3.2. Improved Spatial Pyramid Pooling in YOLOv2

The multi-scale prediction of YOLOv2 and YOLOv3 focuses on concatenating the global features of multi-scale convolutional layers while ignores the fusion of multi-scale local region features on the same convolutional layer. Moreover, although the FPN structure introduced by YOLOv3 improves the accuracy of small target detection, it also has comparatively worse performance on medium and larger size objects [22].

Consequently, a new space pyramid pooling block was designed and introduced into YOLOv2 for pooling and concatenating the multi-scale local region features. The global and local multi-scale features are utilized together to improve the accuracy of object detection.

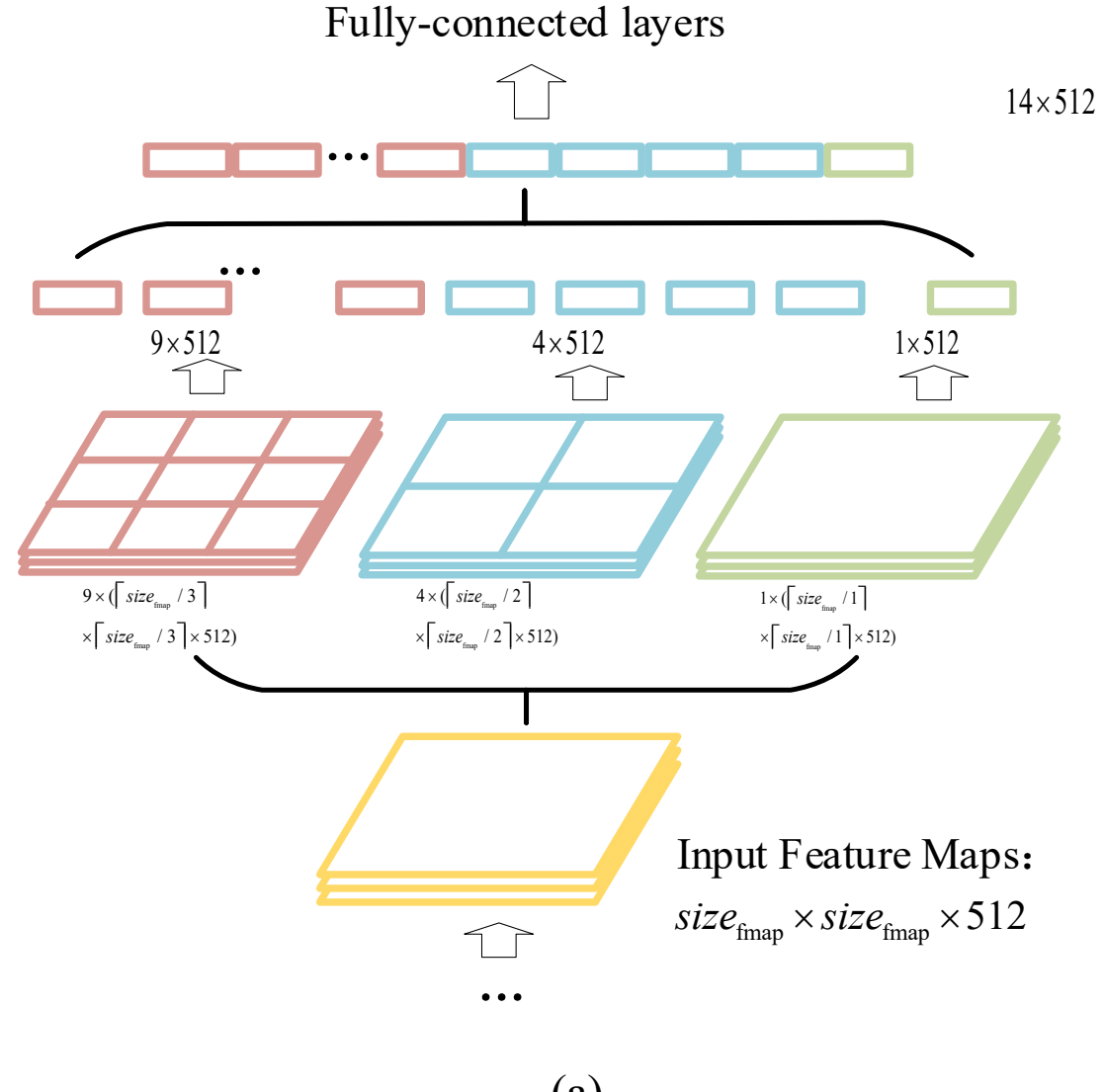

(a)

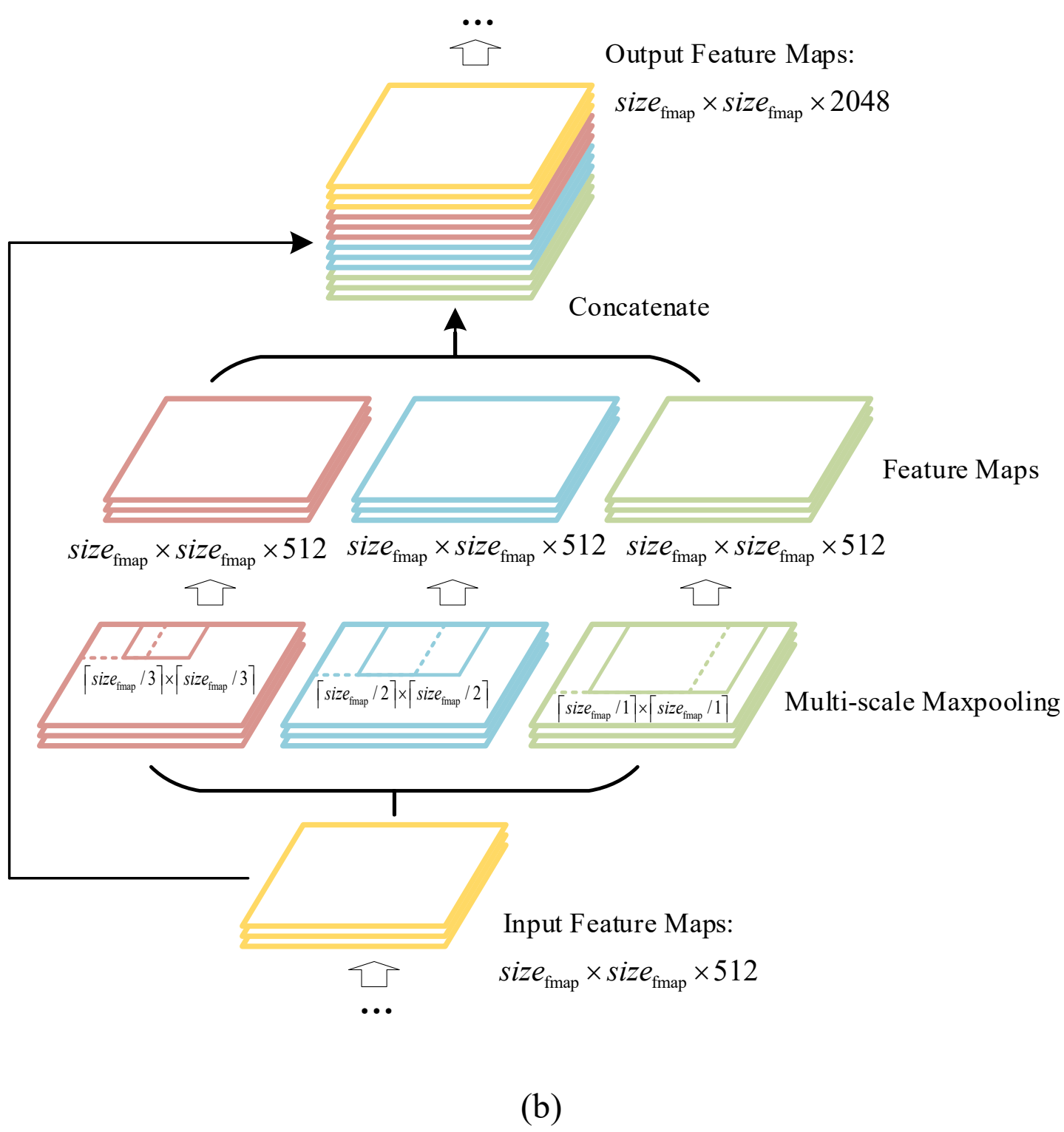

(b)

**Fig. 4.** The Spatial Pyramid Pooling (a) and the Improved Spatial Pyramid Pooling (b).

As shown in Fig. 4(a), the classical spatial pyramid pooling divides the input feature map into $a_i = n_i \times n_i$ bins (where $a_i$ represents the number of bins in the $i$th layer of the feature pyramid) according to the scales that represent different layers of the feature pyramid. The feature maps are pooled by the sliding windows of which the size is the same as that of the bins. The $a_i \cdot d$ -dimensional feature vector where $d$ is the number of filters is obtained to be the input of the fully connected layer.

Our new spatial pyramid pooling block with three max-pooling layers illustrated in Fig. 4(b) is introduced between the DC block and the object detection layer in the network. The 1×1 convolution is utilized to reduce the number of input feature maps from 1024 to 512. After that, the feature maps are pooled in different scales. In Formula (7), $size_{\text{pool}} \times size_{\text{pool}}$ denotes the size of the sliding windows and $size_{\text{fmap}} \times size_{\text{fmap}}$ denotes the size of the feature maps, then

$$size_{\text{pool}} = \left\lceil size_{\text{fmap}} / n_i \right\rceil \tag{7}$$

We let $n_i$= 1, 2, 3 and pool the feature maps by the different sliding windows of which the sizes are $\left\lceil size_{\text{fmap}} / 3 \right\rceil \times \left\lceil size_{\text{fmap}} / 3 \right\rceil$, $\left\lceil size_{\text{fmap}} / 2 \right\rceil \times \left\lceil size_{\text{fmap}} / 2 \right\rceil$ and $\left\lceil size_{\text{fmap}} / 1 \right\rceil \times \left\lceil size_{\text{fmap}} / 1 \right\rceil$ respectively , then the stride of pooling is all 1. The padding is utilized to keep a constant size of the output feature maps, three feature maps with the sizes of $size_{\text{fmap}} \times size_{\text{fmap}} \times 512$ could be obtained.

Different from the traditional spatial pyramid pooling presented by He K et al. [41], our SPP block (Spatial Pyramid Pooling block) in SPP-YOLO (Spatial Pyramid Pooling Based YOLO) does not resize the feature maps into feature vectors with the fixed size. Instead , the three feature maps pooled with the sizes of $size_{\text{fmap}} \times size_{\text{fmap}} \times 512$ and the input feature maps of the SPP block was concatenated to get $size_{\text{fmap}} \times size_{\text{fmap}} \times 2048$ feature maps which extract and converges the multi-scale local region features as the output for object detection.

**Table 2**

The comparison of SPP-YOLO and YOLOv2.

| Method | BFLOP/s | mAP (%) on VOC 2007 | Speed (fps) |
|---|---|---|---|
| YOLOv2 | 29.371 | 76.8 | 67 |
| SPP-YOLO | 29.746 | 77.5 | 64.6 |

The model complexity of YOLOv2 and SPP-YOLO are compared in Table 2 as well as the detection accuracy and speed of YOLOv2 and DC-YOLO on the PASCAL VOC 2007 dataset (see Section 5 for specific experimental settings). As shown in Table 2 that the object detection accuracy of DC-YOLO is 77.5% on the PASCAL VOC 2007 dataset, which is 0.7% higher than that of the YOLOv2. Although the model complexity of DC-YOLO is increased by 0.375 BFLOP/s compared to YOLOv2, the detection speed only reduces by about 2.4 fps. This means that SPP-YOLO improves detection accuracy without significantly increasing the model complexity and reducing the detection speed.

## 3.3. Dense Connection and Spatial Pyramid Pooling Based YOLO

（1）DC-SPP-YOLO Network

The network of DC-SPP-YOLO consisting of five laminated convolution-pooling blocks, a dense connection block with four dense units, a spatial pyramid pooling block with three max-pooling layers and a multi-scale object detection block is shown in Figure 1.

Firstly, the five laminated convolution-pooling blocks decrease the size of features maps' size to 1/32 of the size of input image and increase the number of features maps to 512 by extracting and gathering the image features. After that, the DC block with four dense units composed by 3×3 and 1×1 densely connected convolutional layers, in which the increments of feature maps are set to be 256, 512, 512, and 512 respectively, strengthens the feature extraction and outputs 2304 concatenated feature maps. As a result, the number of output feature maps is reduced by 3×3×1024 filters.

The SPP block with three max-pooling layers is introduced after the DC block for concatenating the local region features extracted and converged by multi-scale pooling. The 1×1 convolution is adopted before the pooling to reduce the number of input feature maps from 1024 to 512. After that, feature maps are pooled by the sliding windows of which the sizes are $\lceil size_{\text{fmap}}/3 \rceil \times \lceil size_{\text{fmap}}/3 \rceil$ , $\lceil size_{\text{fmap}}/2 \rceil \times \lceil size_{\text{fmap}}/2 \rceil$ and $\lceil size_{\text{fmap}}/1 \rceil \times \lceil size_{\text{fmap}}/1 \rceil$ respectively. Then the feature maps pooled and the input feature maps of the SPP block was concatenated to get $size_{\text{fmap}} \times size_{\text{fmap}} \times 2048$ feature maps as the outputs of the SPP block.

The last part of the network is object detection block, in which the output feature maps of DC

block with higher resolution are reconstructed and concatenated with the output feature maps of SPP block with lower resolution. Then the feature maps above are convoluted by the $1\times1\times(K\times(5+C))$ convolution to obtain $S\times S\times(K\times(5+C))$ feature maps for object detection.

**Table 3**

The network's parameters of DC-SPP-YOLO

| Layers | Parameters | | Output | Layers | Parameters | | Output |
|---|---|---|---|---|---|---|---|
| | Filters | Size / Stride | | | Filters | Size / Stride | |
| Conv 1 | 32 | 3×3/ 1 | 416×416×32 | **DC Block** | **1024** | **3×3 / 1** ×4 | **13×13×2304** |
| Maxpool 1 | | 2×2 / 2 | 208×208×32 | **Conv 14-21** | **256 or 512** | **1×1 / 1** | |
| Conv 2 | 64 | 3×3 / 1 | 208×208×64 | **Conv 22-25** | **1024** | **3×3 / 1** ×2 | **13×13×1024** |
| Maxpool 2 | | 2×2 / 2 | 104×104×64 | | **512** | **1×1 / 1** | **13×13×512** |
| Conv 3 | 128 | 3×3 / 1 | 104×104×128 | **SPP Block Maxpool 6-8** | | **5×5 / 1** **Concat** | **13×13×2048** |
| Conv 4 | 64 | 1×1 / 1 | 104×104×64 | | | **7×7 / 1** | |
| Conv 5 | 128 | 3×3 / 1 | 104×104×128 | | | **13×13 / 1** | |
| Maxpool 3 | | 2×2 / 2 | 52×52×128 | **Conv 26** | **512** | **1×1 / 1** | **13×13×512** |
| Conv 6 | 256 | 3×3 / 1 | 52×52×256 | Conv 27 | 1024 | 3×3 / 1 | 13×13×1024 |
| Conv 7 | 128 | 1×1 / 1 | 52×52×128 | Reorg Conv13 | | / 2 | 13×13×256 |
| Conv 8 | 256 | 3×3 / 1 | 52×52×256 | Concat -1, -2 | | | 13×13×1280 |
| Maxpool 4 | | 2×2 / 2 | 26×26×256 | Conv 28 | 1024 | 3×3 / 1 | 13×13×1024 |
| Conv 9-12 | 512 | 3×3 / 1<br>1×1 / 1 ×2 | Conv 31 | Conv29 | $K$*5+$C$ | 1×1 / 1 | 13×13×($K$*5+$C$) |
| Conv 13 | 512 | 3×3 / 1 | 26×26×512 | Detection | | | |
| Maxpool 5 | | 2×2 / 2 | 13×13×512 | | | | |

Table 3 shows the parameter settings of the DC-SPP-YOLO network and the output of each layer when the size of the input image is 416×416×3. In Table 3, the column of "Layers" represents the different layers of the convolutional neural network; the column of "Filters" represents the channels of the convolution kernel; the column of "Size / Stride" indicates the size / sliding step of the convolution kernel; the column of "Output" represents the size of the output feature map.

（2）Loss Function

The predictions of DC-SPP-YOLO for each bounding box can be denoted as $\boldsymbol{b} = [b_x, b_y, b_w, b_h, b_c]^{\mathrm{T}}$, where $(b_x, b_y)$ is the center coordinates of the box, $b_w$ and $b_h$ are the width and height of the box and $b_c$ is the confidence. The offsets $t_x$, $t_y$ from the top-left corner of the image to the grid center in $b_x$, $b_y$ and the confidence $b_c$ are constrained to [0, 1] by the sigmoid function. Similarly, the ground truth of the bounding box can be denoted as $\boldsymbol{g} = [g_x, g_y, g_w, g_h, g_c]^{\mathrm{T}}$. The classification result of each bounding box is $\boldsymbol{Class} = [Class_1, Class_2, \ldots, Calss_C]^{\mathrm{T}}$, then the ground truth of the classification is $\Pr(Class_l)_{l\in C}$, and the predicted probability that the object belongs to the $l$ class is $\widehat{\Pr}(Class_l)_{l\in C}$.

In this paper, a new loss function is constructed for the CNN model training, which applied the mean squared error of the coordinate regression and the cross-entropy of object classification to describe the loss of object detection. Compared with only using the mean squared error to represent both of the coordinate regression loss and the object classification loss in YOLOv2, using the cross-entropy to represent the object classification loss can alleviate the vanishing-gradient and make the model training robust. The new loss function constructed is shown in Eq. 8.

$$
\begin{aligned}
\boldsymbol{L}(\boldsymbol{b},\boldsymbol{Class}) = {} & \lambda_{\text{noobj}}\sum_{i=0}^{S}\sum_{j=0}^{S}\sum_{k=0}^{K}1_{ijk}^{\text{noobj}}\cdot\left((g_{c^{ij}}-b_{c^{ijk}})\cdot\nabla_{\sigma}(b_{c^{ijk}})\right)^{2} \\
& +\lambda_{\text{obj}}\sum_{i=0}^{S}\sum_{j=0}^{S}\sum_{k=0}^{K}1_{ijk}^{\text{obj}}\cdot\left((g_{c^{ij}}-b_{c^{ijk}})\cdot\nabla_{\sigma}(b_{c^{ijk}})\right)^{2} \\
& +\lambda_{\text{coord}}\sum_{i=0}^{S}\sum_{j=0}^{S}\sum_{k=0}^{K}1_{ijk}^{\text{obj}}\cdot\Big(\left((g_{x^{ij}}-b_{x^{ijk}})\cdot\nabla_{\sigma}(b_{x^{ijk}})\right)^{2} \\
& \qquad +\left((g_{y^{ij}}-b_{y^{ijk}})\cdot\nabla_{\sigma}(b_{y^{ijk}})\right)^{2} \\
& \qquad +(g_{w^{ij}}-b_{w^{ijk}})^{2}+(g_{h^{ij}}-b_{h^{ijk}})^{2}\Big) \\
& +\lambda_{\text{class}}\sum_{i=0}^{S}\sum_{j=0}^{S}\sum_{k=0}^{K}1_{ijk}^{\text{obj}}\sum_{l=1}^{C}\left(-\Pr_{ij}(Class_{l})\log(\widehat{\Pr}_{ijk}(Class_{l}))\right) \\
& +\lambda_{\text{prior}}\sum_{i=0}^{S}\sum_{j=0}^{S}\sum_{k=0}^{K}1_{ijk}^{\text{prior}}\cdot\Big(\left((prior_{x^{ijk}}-b_{x^{ijk}})\cdot\nabla_{\sigma}(b_{x^{ijk}})\right)^{2} \\
& \qquad +\left((prior_{y^{ijk}}-b_{y^{ijk}})\cdot\nabla_{\sigma}(b_{y^{ijk}})\right)^{2} \\
& \qquad +(prior_{w^{ijk}}-b_{w^{ijk}})^{2}+(prior_{h^{ijk}}-b_{h^{ijk}})^{2}\Big)
\end{aligned}
\tag{8}
$$

In the loss function above, if the maximum of $\text{IoU}_{\text{pred}}^{\text{truth}}$ is greater than the threshold $\text{IoU}_{\text{thres}}$, $1_{ijk}^{\text{obj}}=1,\ 1_{ijk}^{\text{noobj}}=0$; otherwise $1_{ijk}^{\text{obj}}=0,\ 1_{ijk}^{\text{noobj}}=1$. The gradient of the sigmoid function is $\nabla_{\sigma}(.)$. Since only the maximum of $\text{IoU}_{\text{pred}}^{\text{truth}}$ is taken as the prediction result of each grid among the $K$ anchor boxes, without considering the loss of the other anchor boxes may result in instability of model training. In order to improve the stability of model training and make the model easily learn the shape of object, we calculate the loss between these bounding boxes above and those bounding boxes which do not provide useful predictions. When the number of trained samples is less than $N_{\text{prior}}$, $1_{ijk}^{\text{prior}}=1$, the predictions of the prior box can be represented as ***Prior*** = [$Prior_x$, $Prior_y$, $Prior_w$, $Prior_h$]$^{\text{T}}$; otherwise, $1_{ijk}^{\text{prior}}=0$. Besides, the hyperparameters $\lambda_{\text{noobj}}$, $\lambda_{\text{obj}}$, $\lambda_{\text{coord}}$, $\lambda_{\text{class}}$ and $\lambda_{\text{prior}}$ are the weight coefficients on each part of the loss function respectively.

**Table 4**

The comparison of YOLOv2 and YOLOv2 using new loss function.

| Method | mAP (%) on VOC 2007 | Convergence (epoch) |
|---|---|---|
| YOLOv2 | 76.8 | 160 |
| YOLOv2 + new loss function | 77.0 | 145 |

The object detection accuracy of YOLOv2 and YOLOv2 using new loss function on the PASCAL VOC 2007 dataset are compared in Table 4. It's shows that the object detection accuracy of YOLOv2 using new loss function is 77.0% on the PASCAL VOC 2007 dataset, which is 0.2% higher than that of the YOLOv2. In the same environment, we also campare the convergent rate of two methods above. As shown in Table 4, the YOLOv2 network model converged at the 160th epoch while the network model of YOLOv2 using new loss function converged at the 145th epoch when training. This means that the new loss function can not only improve detection accuracy but make the network model robust and converge faster when training.

## 4. Object Detection Using DC-SPP-YOLO

The object detection process based on DC-SPP-YOLO which includes dataset construction, model training, and object detection is shown in Fig. 5.

Firstly, data augmentation methods such as random crop, scale augmentation, PCA jittering, are used to preprocess the training images for improving object detection performance and preventing the model from over-fitting. The k-means clustering is run for anchor boxes generation instead of hand-picked priors, and the IoU (Intersection-over-Union) between the bounding boxes of training samples and clustering centroids is utilized for constructing the distance metric

$$dist_{\text{centroid}}^{\text{box}} = 1 - \text{IoU}_{\text{centroid}}^{\text{box}} \tag{9}$$

Then, the training parameters are set and the convolutional neural network is loaded. The loss function is constructed with the sum of squared errors loss on regression and the binary cross-entropy loss on classification. The weights of the model are updated iteratively to make the loss function converge, and the DC-SPP-YOLO model is obtained for object detection.

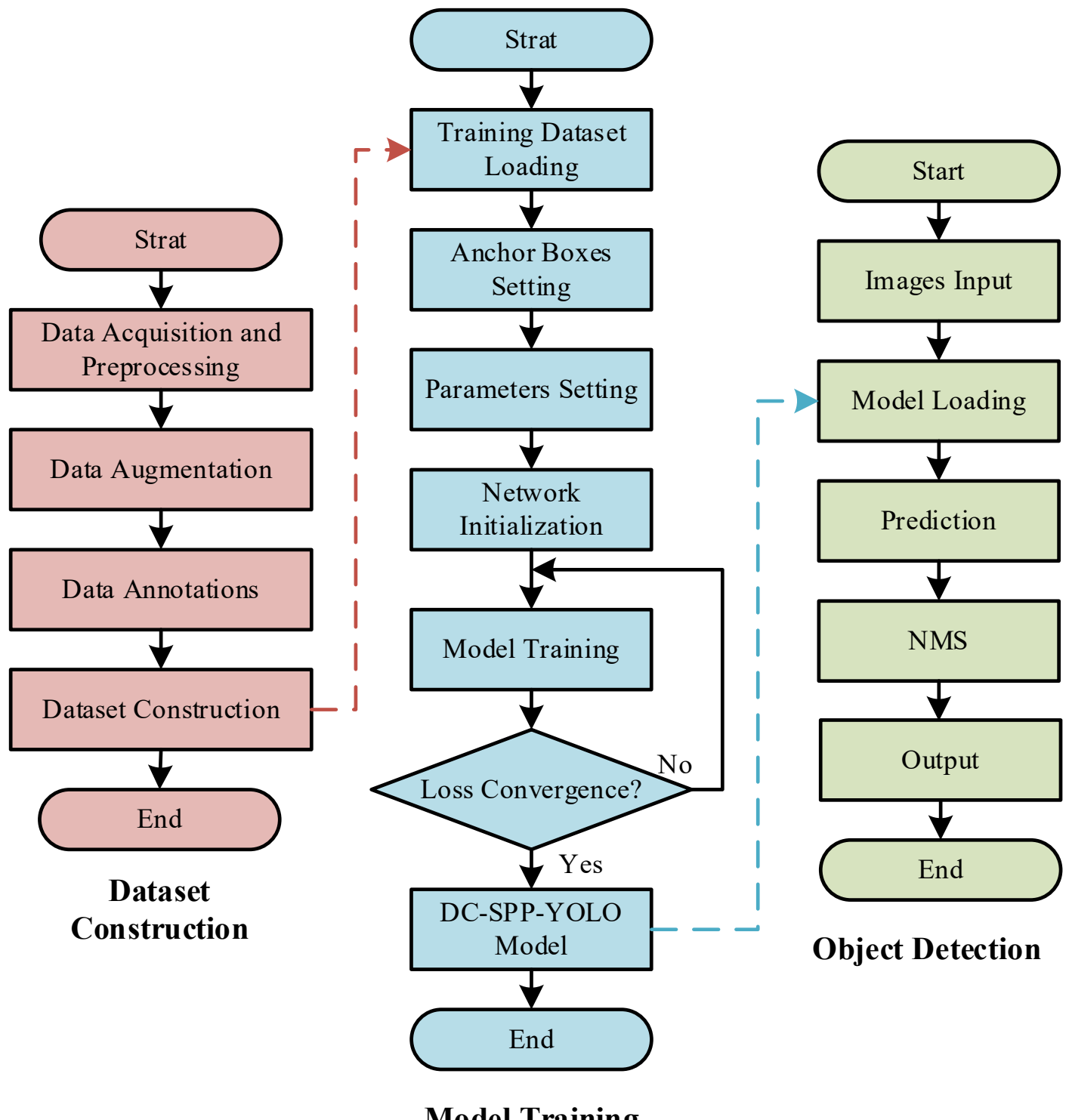

**Fig. 5.** The object detection process based on DC-SPP-YOLO.

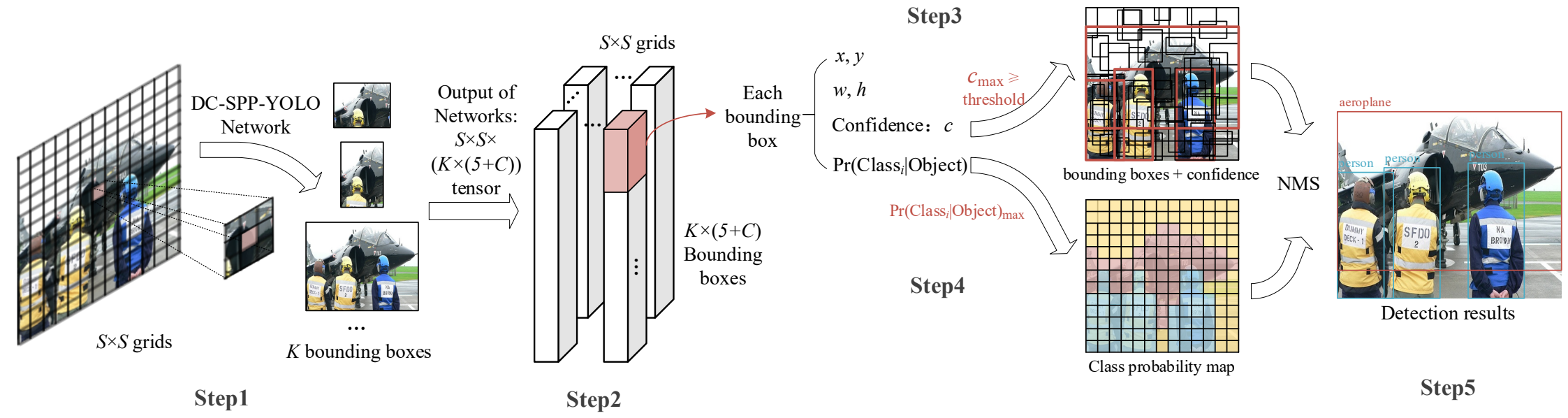

**Fig. 6.** The flow chart of object detection using DC-SPP-YOLO.

The algorithm's flowchart of DC-SPP-YOLO for object detection is shown in Fig. 6.

Step1: Divide the input image into $S$×$S$ grids; each grid generates $K$ bounding boxes according to the anchor boxes.

Step2: Use the convolutional neural network to extract object features and predict the $\boldsymbol{b} = [b_x, b_y, b_w, b_h, b_c]^{\mathrm{T}}$ and the $\boldsymbol{Class} = [Class_1, Class_2, \ldots, Calss_C]^{\mathrm{T}}$.

Step3: Compare the maximum confidence $\mathrm{IoU}_{\mathrm{pred}}^{\mathrm{truth}}$ of the $K$ bounding boxes with the threshold $\mathrm{IoU}_{\mathrm{thres}}$;

if $\mathrm{IoU}_{\mathrm{pred}}^{\mathrm{truth}} > \mathrm{IoU}_{\mathrm{thres}}$, the bounding box contains the object;

else, the bounding box does not contain the object.

Step4: Choose the category with the highest predicted probability as the object category.

Step5: Adopt NMS (Non-Maximum Suppression) to perform a maximum local search for suppressing redundant boxes, output and display the results of object detection.

# 5. Experiments

## 5.1. Experimental Setup and Implementation

（1）Experimental Condition

Experiments were run on a Windows 10 PC with an Intel Xeon E5-2643 3.3GHz CPU, 32GB memory and an NVIDIA Titan X GPU with 12.00 GB memory. The program was developed on the Visual Studio 2017 platform by C/C++ language, and the deep learning framework Darknet was used.

（2）Datasets

The effectiveness of DC-SPP-YOLO compared with state-of-the-art methods was demonstrated, especially with YOLOv2, on PASCAL VOC dataset and UA-DETRAC dataset.

The PASCAL VOC datasets used for the object detection experiments were set as follows. The experimental datasets contained 32487 images in which the objects belonged to twenty categories. The VOC 2007 trainval dataset and the VOC 2012 trainval dataset were used to train the DC-SPP-YOLO model. The VOC 2007 test dataset and the VOC 2012 test dataset were used to test the performance of DC-SPP-YOLO. As with the object detection methods such as YOLOv2, the $\mathrm{IoU}_{\mathrm{pred\ thres}}^{\mathrm{truth}}$ was set to 0.5. The result on the PASCAL VOC 2012 test dataset was given by PASCAL VOC Challenge Evaluation Server.

The UA-DETRAC dataset used for the object detection experiments contained 82088 vehicle images taken by traffic cameras, in which the objects belonged to four categories. There were 20522 images for model training, 20522 images for validation and 41044 images for the test.

（3）Model Training

The training parameters of DC-SPP-YOLO were set as follow:

① The parameter $a_i$ of leak ReLU was 10, which is the same as YOLOv2.

② Our loss function has a similar structure to that of YOLOv2. For ensuring a comparative analysis of the loss function under the same experimental conditions, we chose the same hyperparameters values. So, the hyperparameters $\lambda_{\text{noobj}}$, $\lambda_{\text{obj}}$, $\lambda_{\text{coord}}$, $\lambda_{\text{class}}$ and $\lambda_{\text{prior}}$ of our loss function were 1, 5, 1, 1 and 0.1 respectively. The $N_{\text{prior}}$ was 12800 in our experimental experience.

③ The adaptive moment estimation (Adam) was used to update the weights of the network [45], where the momentum was 0.9, and the decay was 0.0005. According to the GPU memory used in the experiments, the batch size was set as 64.

④ We trained the network for 145 epochs with a starting learning rate of 0.001 (according to the experimental environment and datasets), and the learning rate on the 20th epoch and the 70th epoch was reduced to 0.1 times of the original.

（4）Evaluation

The object detection accuracy was measured by mAP (mean Average Precision) when $\text{IoU}_{\text{thres}} = 0.5$, and the detection speed was represented by fps (frames per second).

## 5.2. Experiments on PASCAL VOC 2007

The detailed experimental results of DC-SPP-YOLO on the PASCAL VOC 2007 test dataset was shown in Table 5. The detection accuracy and speed of DC-SPP-YOLO with the state-of-the-art methods was compared in Table 6. When the size of the input image is 416×416 pixels, the DC-SPP-YOLO is represented as DC-SPP-YOLO 416, and other methods are also represented as described above.

Since Redmon J and Farhadi A [20] did not give the test results of YOLOv3 on the PASCAL VOC dataset, we used their open source code from *https://pjreddie.com/darknet/yolo/* for object detection experiments on the PASCAL VOC 2007 test dataset. Then, the experimental results of YOLOv3 was used as a control group for the experimental results of DC-SPP-YOLO.

As shown in Table 4 and Table 5, at 55.7 fps, the mAP of DC-SPP-YOLO 416 is 78.4%, which is 1.6% higher than that of YOLOv2 416. At 37.9 fps, the mAP of DC-SPP-YOLO 544 is 79.6%, which is 1.0% higher than that of YOLOv2 544; the accuracy improvement above only slightly decreases the detection speed. Although the mAP of YOLOv3 416 is 79.3% in the experiments, the speed of YOLOv3, which is as fast as DC-SPP-YOLO 544 but lower than DC-SPP-YOLO 416 and YOLOv2 416, has been damaged due to the larger backbone network

Darknet53 with residual connection structure.

**Table 5**

The detection results of DC-SPP-YOLO on PASCAL VOC2007 test dataset.

| Method | aero | bike | bird | boat | bottle | bus | car | cat | chair | cow | table | dog | horse | mbike | person | plant | sheep | sofa | train | tv | mAP (%) | Speed (fps) |
|---|---|---|---|---|---|---|---|---|---|---|---|---|---|---|---|---|---|---|---|---|---|---|
| DC-SPP-YOLO 416 | 80.0 | 84.9 | 76.0 | 68.0 | 53.8 | 87.6 | 83.9 | 90.1 | 62.5 | 84.1 | 75.8 | 88.6 | 87.3 | 85.7 | 77.0 | 54.3 | 81.7 | 80.1 | 88.3 | 78.7 | 78.4 | 56.3 |
| DC-SPP-YOLO 544 | 83.1 | 85.9 | 77.2 | 69.5 | 59.7 | 88.5 | 86.3 | 89.9 | 62.6 | 86.0 | 78.3 | 87.6 | 88.0 | 86.7 | 80.1 | 54.3 | 81.3 | 80.4 | 87.6 | 79.4 | 79.6 | 38.9 |

**Table 6**

The comparison of accuracy and speed on PASCAL VOC2007 test dataset

| Method | Year | Base network | mAP (%) | Speed (fps) | GPU |
|---|---|---|---|---|---|
| Faster RCNN[10] | 2015 | VGG16 | 73.2 | 7.0 | Titan X |
| Faster RCNN[14] | 2016 | ResNet-101 | 76.4 | 5.0 | Titan X |
| SSD 300[17] | 2016 | VGG16 | 74.3 | 46.0 | Titan X |
| SSD 512[17] | 2016 | VGG16 | 76.8 | 19.0 | Titan X |
| DSSD 321[19] | 2017 | ResNet-101 | 78.6 | 9.5 | Titan X |
| DSSD 513[19]] | 2017 | ResNet-101 | 81.5 | 5.5 | Titan X |
| STDN 300[21] | 2018 | DenseNet-169 | 78.1 | 40.3 | Titan X |
| STDN 513[21] | 2018 | DenseNet-169 | 80.9 | 28.1 | Titan X |
| YOLO[16] | 2016 | Darknet19 | 63.4 | 45.0 | Titan X |
| YOLOv2 416[18] | 2017 | Darknet19 | 76.8 | 67.0 | Titan X |
| YOLOv2 544[18] | 2017 | Darknet19 | 78.6 | 40.0 | Titan X |
| YOLOv3 416[20] | 2018 | Darknet53 | 79.3 | 37.9 | Titan X |
| DC-SPP-YOLO 416 | 2018 | Darknet29 | 78.4 | 55.7 | Titan X |
| DC-SPP-YOLO 544 | 2018 | Darknet29 | 79.6 | 38.4 | Titan X |
| YOLOv3(DC) 416 | 2018 | Darknet54 | 79.5 | 37.7 | Titan X |
| YOLOv3(DC)+SPP 416 | 2018 | Darknet54 | 79.7 | 37.0 | Titan X |

To further test the proposed method, we also compare the DC-SPP-YOLO method with YOLOv3, which also improves the backbone network and multi-scale detection of YOLOv2. The DC structure proposed was completely used instead of the laminated convolution-pooling structure of YOLOv2. Then, the YOLOv3(DC) with 54 convolutional layers in the backbone network was obtained as a control group. The SPP block proposed was introduced based on YOLOv3(DC), and the YOLOv3(DC)+SPP was obtained as a control group. The experimental results of YOLOv3, YOLOv3(DC) and YOLOv3(DC)+SPP also verified the effectiveness of the proposed method.

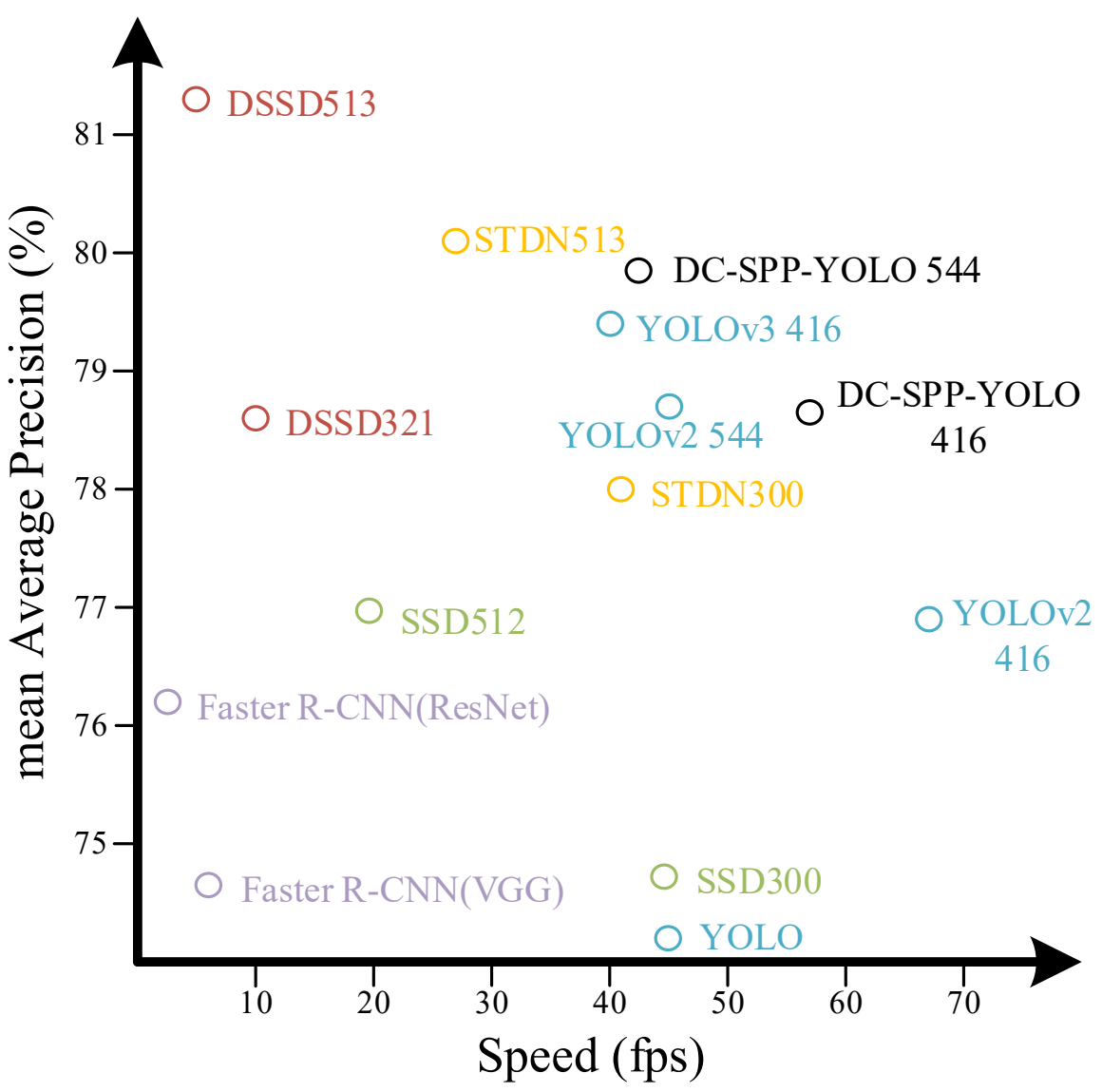

**Fig. 7.** The accuracy and speed on PASCAL VOC2007.

Detection performance of DC-SPP-YOLO was compared with the state-of-the-art methods in Table 6. The performance of DC-SPP-YOLO is obviously better than that of Faster R-CNN and YOLO on Pascal VOC 2007 test dataset. The DC-SPP-YOLO 544 is not only more accurate than SSD 512 and but also runs twice as fast as SSD 512. Even though the mAP of DC-SPP-YOLO 544 is 1.9% lower than that of DSSD 513, the detection speed is much faster than that of DSSD 513 (more than seven times ) due to the severe constrainment of the detection speed of the DSSD is by the extremely deep backbone network (ResNet-101) and the inefficient feature fusion. STDN 513 adopts the DenseNet-169 backbone network to improve the speed of DSSD 513, but STDN 513 is still about one third slower compare to DC-SPP-YOLO 544 even though STDN 513 has a 1.3% higher mAP compare to DC-SPP-YOLO 544. As shown in Fig. 7, considering both the detection accuracy and speed, the general performance of our method is better than that of STDN.

**Table 7**

The path from YOLOv2 to DC-SPP-YOLO.

| Improvement | YOLOv2 416 | | | | | DC-SPP-YOLO 416 |
|---|---|---|---|---|---|---|
| Dense Connection | | √ | | | √ | √ |
| Spatial Pyramid Pooling | | | √ | | √ | √ |
| New loss function | | | | √ | | √ |
| mAP (%) | 76.8 | 77.6 | 77.5 | 77.0 | 78.0 | 78.4 |
| Speed (fps) | 67.0 | 58.9 | 64.6 | 67.0 | 55.7 | 55.7 |

The comparison of the improvement on each component in DC-SPP-YOLO is shown in Table 7. The mAP of DC-YOLO in which the improved dense connection of convolutional layers is employed is 0.8% higher than that of YOLOv2. The mAP of SPP-YOLO in which the improved spatial pyramid pooling block is introduced is 0.7% higher than that of YOLOv2. The mAP of YOLOv2 using new loss function is 0.2% higher than that of YOLOv2. And the DC-SPP-YOLO without using new loss function gets a 1.2% higher mAP than that of YOLOv2, while the mAP of DC-SPP-YOLO using new loss function is 1.6% higher than that of YOLOv2 which shows the new loss function make a 0.4% higher mAP contribution on DC-SPP-YOLO. According to the results of these experiments, the mAP of DC-SPP-YOLO using new loss function was improved from 76.8% to 78.4%, which further demonstrates the effectiveness of the above three improved methods respectively.

## 5.3. Experiments on PASCAL VOC 2012

Results of the Object detection experiments on the PASCAL VOC 2012 dataset were shown in Table 8 and Fig. 8, which demonstrated the good performance of DC-SPP-YOLO further. The experimental results have been evaluated by the PASCAL VOC Evaluation Server. The evaluation results can be found at *http://host.robots.ox.ac.uk:8080/anonymous/TAD5II.html*. At 38.4 fps, the mAP of DC-SPP-YOLO 544 is 1.2% higher than that of YOLOv2 544. Moreover, the results show that the AP variation from DC-SPP-YOLO to YOLOv2 is much higher in the categories of "sheep/car" (more small-sized objects) and "bus/mbike/table" (more large-sized objects) than that in the other categories. It demonstrates the detection accuracy of YOLOv2 is ameliorated by the improved dense connection and the improved spatial pyramid pooling.

As shown in Table 9, the object detection performance of the DC-SPP-YOLO was compared with the state-of-the-art methods on the PASCAL VOC 2012 dataset. The experimental results on PASCAL VOC 2012 are similar to those on PASCAL VOC 2007. The mAP of DC-SPP-YOLO 544 is higher than that of Faster R-CNN and YOLOv2, and is comparable to that of SSD 512, but is lower than that of DSSD. On the other side, DC-SPP-YOLO 544 runs much faster than Faster R-CNN, DSSD 513 and SSD 512, only runs a little slower than YOLOv2. In general, these results above demonstrate the effectiveness of DC-SPP-YOLO both in detection accuracy and speed further.

**Table 8**

The detection results of DC-SPP-YOLO on PASCAL VOC2012 test dataset.

| Method | aero | bike | bird | boat | bottle | bus | car | cat | chair | cow | table | dog | horse | mbike | person | plant | sheep | sofa | train | tv | mAP (%) |
|---|---|---|---|---|---|---|---|---|---|---|---|---|---|---|---|---|---|---|---|---|---|
| YOLOv2 544[18] | 86.3 | 82.0 | 74.8 | 59.2 | 51.8 | 79.8 | 76.5 | 90.6 | 52.1 | 78.2 | 58.5 | 89.3 | 82.5 | 83.4 | 81.3 | 49.1 | 77.2 | 62.4 | 83.8 | 68.7 | 73.4 |
| DC-SPP-YOLO 544 | 86.9 | 82.5 | 75.7 | 60.1 | 52.9 | 82.5 | 78.4 | 91.0 | 52.8 | 80.2 | 60.8 | 89.4 | 83.5 | 85.5 | 82.5 | 49.5 | 79.8 | 63.9 | 83.7 | 68.3 | 74.6 |
| Variation | +0.6 | +0.5 | +0.9 | +0.9 | +1.1 | +2.7 | +1.9 | +0.4 | +0.7 | +2 | +2.3 | +0.1 | +1 | +2.1 | +1.2 | +0.4 | +2.6 | +1.5 | -0.1 | -0.4 | +1.2 |

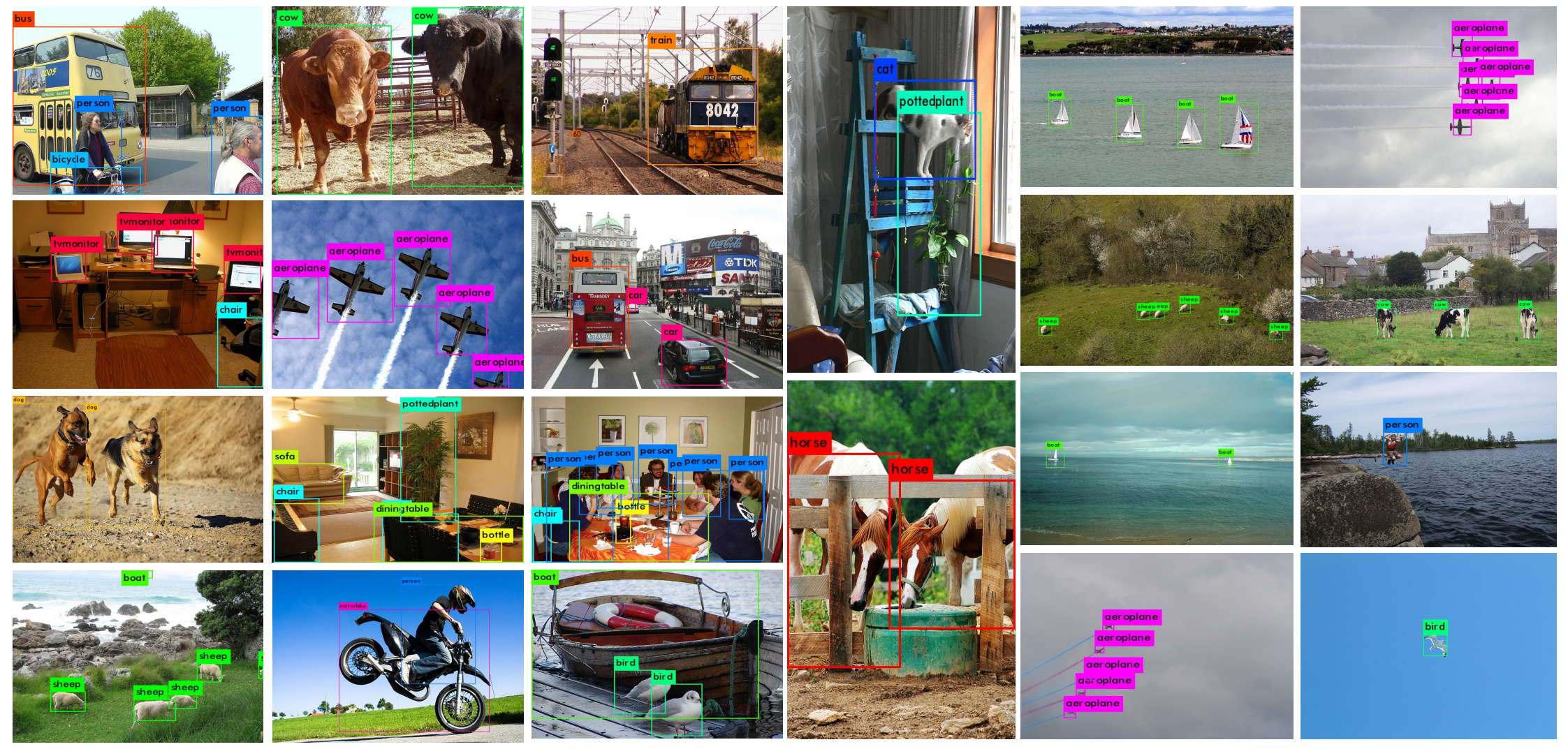

**Fig. 8.** The object detection using DC-SPP-YOLO on PASCAL VOC2012 test dataset.

**Table 9**

The comparison of accuracy and speed on PASCAL VOC2012 test dataset.

| Method | Year | Base network | mAP(%) | Speed(fps) | GPU |
|---|---|---|---|---|---|
| Faster RCNN[10] | 2015 | VGG16 | 70.4 | 7 | Titan X |
| Faster RCNN[14] | 2016 | ResNet-101 | 73.8 | 5 | Titan X |
| SSD 300[17] | 2016 | VGG16 | 72.4 | 46 | Titan X |
| SSD 512[17] | 2016 | VGG16 | 74.9 | 19 | Titan X |
| DSSD 321[19] | 2017 | ResNet-101 | 76.3 | 9.5 | Titan X |
| DSSD 513[19] | 2017 | ResNet-101 | 80.0 | 5.5 | Titan X |
| YOLO[16] | 2016 | Darknet19 | 57.9 | 45 | Titan X |
| YOLOv2 544[18] | 2017 | Darknet19 | 73.4 | 40 | Titan X |
| DC-SPP-YOLO 544 | 2018 | Darknet31 | 74.6 | 38.4 | Titan X |

## 5.4. Experiments for Vehicles Detection on UA-DETRAC

Vehicle detection is one of the significant applications of vision-based object detection methods. In order to verify the detection performance of our method in the real scene, the UA-DETRAC Trainval dataset was utilized to train the DC-SPP-YOLO 416 model and the YOLOv2 416 model. The results of vehicle detection in various environmental conditions are summarized in Table 10. The DC-SPP-YOLO 416 has a 2.25% higher mAP than YOLOv2 416 at 57.5 fps. It indicated that the improved dense connection and the new spatial pyramid pooling are helpful to increase the accuracy.

Comparing with the other methods in Table 9, the DC-SPP-YOLO 416 gets the higher mAP than SSDR, FRCNN-Res and DFCN. Even if the mAP of DC-SPP-YOLO 416 is lower than that of EB, GP-FRCN and RCNN-SC, the DC-SPP-YOLO 416 runs much faster than these methods above. Besides，as shown in Fig. 9, the object detection of DC-SPP-YOLO is robust in complex scenes such as variable lighting conditions, different weather, object occlusion, object blurring.

**Table 10**

The comparison of accuracy and speed on UA-DETRAC dataset.

| Method | mAP(%) | Speed(fps) | GPU |
|---|---|---|---|
| GP-FRCNN[46] | 91.90 | 4.0 | Tesla K40 |
| EB[47] | 89.57 | 11.0 | Titan X |
| SSDR[48] | 79.47 | 34.0 | GTX 1080 |
| RCNN-SC[48] | 93.43 | 2.2 | 2×Tesla K80 |
| FRCNN-Res[48] | 82.90 | 0.6 | Titan X |
| DFCN[48] | 86.86 | 11.0 | Titan X |
| YOLOv2 416[18] | 85.48 | 65.8 | Titan X |
| DC-SPP-YOLO 416 | 87.73 | 57.5 | Titan X |

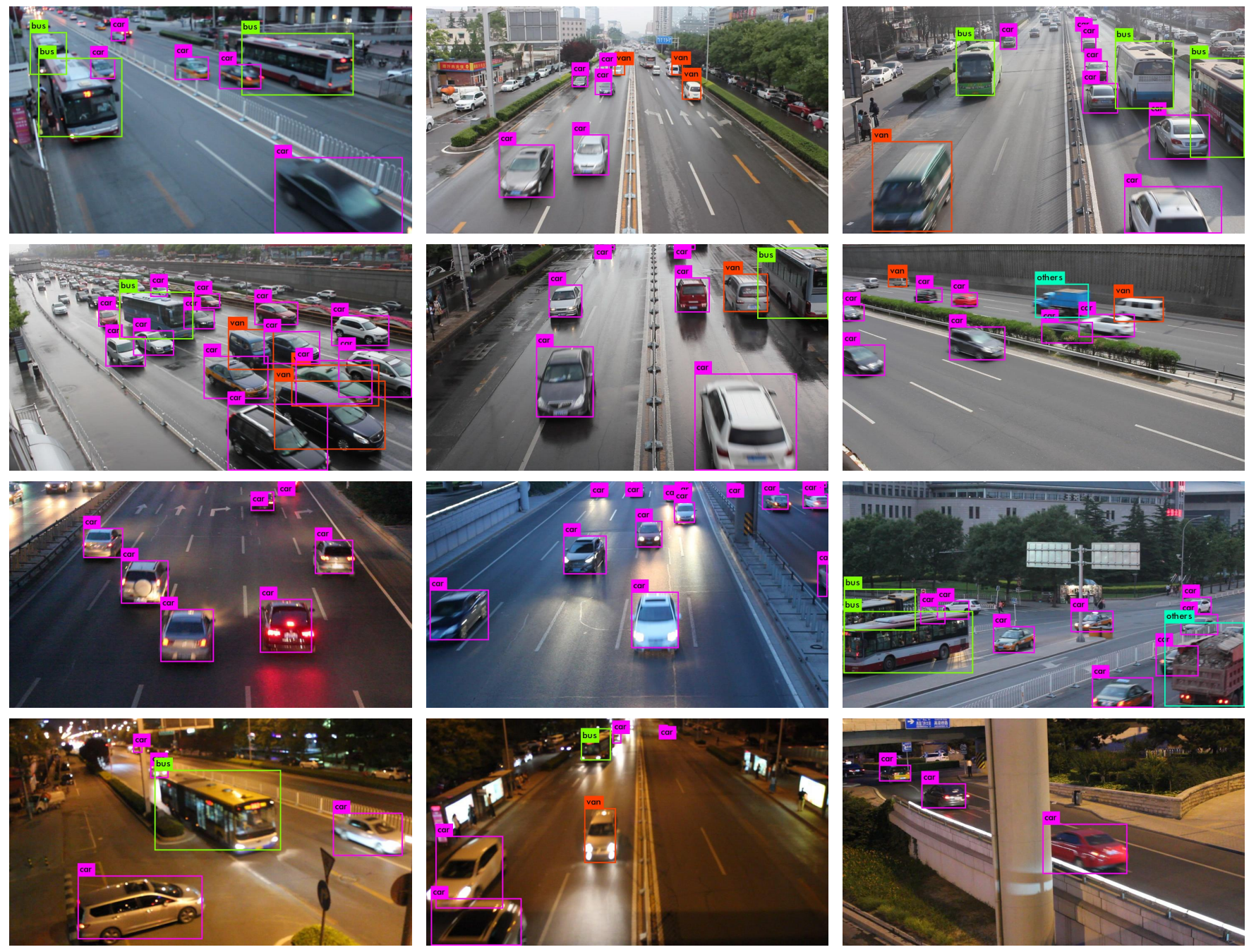

**Fig. 9.** The object detection using DC-SPP-YOLO on UA-DETRAC dataset.

# 6. Conclusions

In this paper, a DC-SPP-YOLO object detection method is proposed for improving the detection accuracy of YOLOv2 while keeping the real-time detection speed. In DC-SPP-YOLO, the improved dense connection structure of the convolutional layers is utilized to strengthen the feature extraction of backbone network in YOLOv2 and to alleviate the vanishing-gradient problem. A improved spatial pyramid pooling structure is introduced to pool the multi-scale region features on different scales in the same convolutional layer. A new loss function is adopted to accelerate the model training. The experiments on the PASCAL VOC datasets and the UA-DETRAC datasets demonstrate that the detection accuracy of DC-SPP-YOLO is higher than that of YOLOv2 and is as good as that of popular methods on the object detection tasks.

In the field of object detection, complex environments, large-scale variances, and rotational variations still constrain the improvement of accuracy. In the future, the research on robust object detection with rotation invariance and scale invariance will be one of the important research contents in this field.

## Acknowledgments

This work is supported by the National Key R&D Program of China (No. 2017YFF0107303).

## References

[1] S. Agarwal, J. O. D. Terrail, and F. Jurie, "Recent Advances in Object Detection in the Age of Deep Convolutional Neural Networks," *ArXiv180903193 Cs*, Sep. 2018.

[2] P. F. Felzenszwalb, R. B. Girshick, D. McAllester, and D. Ramanan, "Object Detection with Discriminatively Trained Part-Based Models," *IEEE Trans. Pattern Anal. Mach. Intell.*, vol. 32, no. 9, pp. 1627–1645, Sep. 2010.

[3] J. Zhang, K. Huang, Y. Yu, and T. Tan, "Boosted local structured HOG-LBP for object localization," in *2011 IEEE Computer Vision and Pattern Recognition (CVPR)*, Colorado Springs, CO, USA, 2011, pp. 1393–1400.

[4] J. Zhang, Y. Huang, K. Huang, Z. Wu, and T. Tan, "Data decomposition and spatial mixture modeling for part based model," in *Asian Conference on Computer Vision*, 2012, pp. 123–137.

[5] A. Krizhevsky, I. Sutskever, and G. E. Hinton, "ImageNet Classification with Deep Convolutional Neural Networks," in *International Conference on Neural Information Processing Systems*, 2012, pp. 1097–1105.

[6] Y. LeCun, Y. Bengio, and G. Hinton, "Deep learning," *Nature*, vol. 521, no. 7553, pp. 436–444, May 2015.

[7] R. Girshick, J. Donahue, T. Darrell, and J. Malik, "Rich Feature Hierarchies for Accurate Object Detection and Semantic Segmentation," in *2014 IEEE Conference on Computer Vision and Pattern Recognition*, Columbus, OH, USA, 2014, pp. 580–587.

[8] J. Han, D. Zhang, G. Cheng, N. Liu, and D. Xu, "Advanced Deep-Learning Techniques for Salient and Category-Specific Object Detection: A Survey," *IEEE Signal Process. Mag.*, vol. 35, no. 1, pp. 84–100, Jan. 2018.

[9] R. B. Girshick, "Fast R-CNN," in *2015 International Conference on Computer Vision (ICCV)*, Santiago, Chile, 2015, pp. 1440–1448.

[10] S. Ren, K. He, R. Girshick, and J. Sun, "Faster R-CNN: Towards Real-Time Object Detection with Region Proposal Networks," *IEEE Trans. Pattern Anal. Mach. Intell.*, vol. 39, no. 6, pp. 1137–1149, Jun. 2017.

[11] J. Dai, Y. Li, K. He, and J. Sun, "R-FCN: Object Detection via Region-based Fully Convolutional Networks," *Neural Inf. Process. Syst.*, pp. 379–387, 2016.

[12] J. Ma *et al.*, "Arbitrary-Oriented Scene Text Detection via Rotation Proposals," *IEEE Trans. Multimed.*, vol. PP, no. 99, pp. 1–1, 2017.

[13] B. Singh, H. Li, A. Sharma, and L. S. Davis, "R-FCN-3000 at 30fps: Decoupling Detection and Classification," in *2018 IEEE Conference on Computer Vision and Pattern Recognition*

*(CVPR)*, Salt Lake City, Utah, USA, 2018, pp. 1081–1090.

[14] K. He, X. Zhang, S. Ren, and J. Sun, “Deep Residual Learning for Image Recognition,” in *2016 IEEE Conference on Computer Vision and Pattern Recognition (CVPR)*, Las Vegas, NV, USA, 2016, pp. 770–778.

[15] Z. Li, Y. Chen, G. Yu, and Y. Deng, “R-FCN++: Towards Accurate Region-based Fully Convolutional Networks for Object Detection,” in *National Conference on Artificial Intelligence*, 2018, pp. 7073–7080.

[16] J. Redmon, S. Divvala, R. Girshick, and A. Farhadi, “You Only Look Once: Unified, Real-Time Object Detection,” in *2016 IEEE Conference on Computer Vision and Pattern Recognition (CVPR)*, Las Vegas, NV, USA, 2016, pp. 779–788.

[17] W. Liu *et al.*, “SSD: Single Shot MultiBox Detector,” in *2016 European Conference on Computer Vision (ECCV)*, Amsterdam, The Netherlands, 2016, vol. 9905, pp. 21–37.

[18] J. Redmon and A. Farhadi, “YOLO9000: Better, Faster, Stronger,” in *2017 IEEE Conference on Computer Vision and Pattern Recognition (CVPR)*, Honolulu, HI, USA, 2017, pp. 6517–6525.

[19] C.-Y. Fu, W. Liu, A. Ranga, A. Tyagi, and A. C. Berg, “DSSD : Deconvolutional Single Shot Detector,” *ArXiv170106659 Cs*, Jan. 2017.

[20] J. Redmon and A. Farhadi, “YOLOv3: An Incremental Improvement,” *ArXiv180402767 Cs*, Apr. 2018.

[21] P. Zhou, B. Ni, C. Geng, J. Hu, and Y. Xu, “Scale-Transferrable Object Detection,” in *2018 IEEE Conference on Computer Vision and Pattern Recognition (CVPR)*, Salt Lake City, Utah, USA, 2018, pp. 528–537.

[22] G. Huang, Z. Liu, L. van der Maaten, and K. Q. Weinberger, “Densely Connected Convolutional Networks,” in *2017 IEEE Conference on Computer Vision and Pattern Recognition (CVPR)*, Honolulu, HI, USA, 2017, pp. 2261–2269.

[23] J. Jeong, H. Park, and N. Kwak, “Enhancement of SSD by concatenating feature maps for object detection,” in *British Machine Vision Conference*, 2017.

[24] K. Lee, J. Choi, J. Jeong, and N. Kwak, “Residual Features and Unified Prediction Network for Single Stage Detection.,” *ArXiv Prepr. ArXiv170705031*, 2017.

[25] G. Cao, X. Xie, W. Yang, Q. Liao, G. Shi, and J. Wu, “Feature-fused SSD: fast detection for small objects,” *ArXiv Comput. Vis. Pattern Recognit.*, vol. 10615, Apr. 2018.

[26] L. Zheng, C. Fu, and Y. Zhao, “Extend the shallow part of single shot multibox detector via convolutional neural network,” in *International Conference on Digital Image Processing*, 2018.

[27] K. Simonyan and A. Zisserman, “Very Deep Convolutional Networks for Large-Scale Image Recognition,” *Int. Conf. Learn. Represent.*, 2015.

[28] R. K. Srivastava, K. Greff, and J. Schmidhuber, “Training very deep networks,” in *Neural Information Processing Systems*, Montreal,Canada, 2015, pp. 2377–2385.

[29] C. Szegedy *et al.*, “Going deeper with convolutions,” in *2016 IEEE Conference on Computer Vision and Pattern Recognition (CVPR)*, Las Vegas, NV, USA, 2015, pp. 1–9.

[30] S. Ioffe and C. Szegedy, “Batch Normalization: Accelerating Deep Network Training by Reducing Internal Covariate Shift,” in *International Conference on Machine Learning*, 2015, pp. 448–456.

[31] C. Szegedy, V. Vanhoucke, S. Ioffe, J. Shlens, and Z. Wojna, “Rethinking the Inception

Architecture for Computer Vision," in *2016 IEEE Conference on Computer Vision and Pattern Recognition (CVPR)*, Las Vegas, NV, USA, 2016, pp. 2818–2826.

[32] C. Szegedy, S. Ioffe, V. Vanhoucke, and A. A. Alemi, "Inception-v4, Inception-ResNet and the Impact of Residual Connections on Learning," in *National Conference on Artificial Intelligence*, 2016, pp. 4278–4284.

[33] F. Chollet, "Xception: Deep Learning with Depthwise Separable Convolutions," in *Computer Vision and Pattern Recognition*, 2017, pp. 1800–1807.

[34] Z. Li, C. Peng, G. Yu, X. Zhang, Y. Deng, and J. Sun, "Light-Head R-CNN: In Defense of Two-Stage Object Detector," *ArXiv Comput. Vis. Pattern Recognit.*, Jan. 2017.

[35] A. G. Howard *et al.*, "MobileNets: Efficient Convolutional Neural Networks for Mobile Vision Applications.," *ArXiv Comput. Vis. Pattern Recognit.*, 2017.

[36] F. N. Iandola, S. Han, M. W. Moskewicz, K. Ashraf, W. J. Dally, and K. Keutzer, "SqueezeNet: AlexNet-level accuracy with 50x fewer parameters and <0.5MB model size," *ArXiv Comput. Vis. Pattern Recognit.*, Apr. 2017.

[37] X. Zhang, X. Zhou, M. Lin, and J. Sun, "ShuffleNet: An Extremely Efficient Convolutional Neural Network for Mobile Devices," in *Computer Vision and Pattern Recognition*, 2018, pp. 6848–6856.

[38] Z. Cai, Q. Fan, R. S. Feris, and N. Vasconcelos, "A Unified Multi-scale Deep Convolutional Neural Network for Fast Object Detection," in *European Conference on Computer Vision*, 2016, pp. 354–370.

[39] F. Yang, W. Choi, and Y. Lin, "Exploit All the Layers: Fast and Accurate CNN Object Detector with Scale Dependent Pooling and Cascaded Rejection Classifiers," in *2016 IEEE Conference on Computer Vision and Pattern Recognition (CVPR)*, 2016, pp. 2129–2137.

[40] J. Li, X. Liang, S. Shen, T. Xu, J. Feng, and S. Yan, "Scale-Aware Fast R-CNN for Pedestrian Detection," *IEEE Trans. Multimed.*, vol. 20, no. 4, pp. 985–996, Apr. 2018.

[41] K. He, X. Zhang, S. Ren, and J. Sun, "Spatial Pyramid Pooling in Deep Convolutional Networks for Visual Recognition," *IEEE Trans. Pattern Anal. Mach. Intell.*, vol. 37, no. 9, pp. 1904–1916, Sep. 2015.

[42] Y. Chen, J. Li, B. Zhou, J. Feng, and S. Yan, "Weaving Multi-scale Context for Single Shot Detector," *ArXiv Comput. Vis. Pattern Recognit.*, Jan. 2017.

[43] T.-Y. Lin, P. Dollar, R. Girshick, K. He, B. Hariharan, and S. Belongie, "Feature Pyramid Networks for Object Detection" in *2017 IEEE Conference on Computer Vision and Pattern Recognition (CVPR)*, Honolulu, HI, 2017, pp. 936–944.

[44] Zeng Dan, et al. "Proposal pyramid networks for fast face detection" in *Information Sciences*, vol. 495, 2019, pp. 136–149.

[45] D. P. Kingma and J. L. Ba, "Adam: A Method for Stochastic Optimization," in *2015 International Conference on Learning Representations (ICLR)*, 2015.

[46] S. Amin and F. Galasso, "Geometric proposals for faster R-CNN" in *Advanced Video and Signal Based Surveillance*, 2017, pp. 1–6.

[47] L. Wang, Y. Lu, H. Wang, Y. Zheng, H. Ye, and X. Xue, "Evolving boxes for fast vehicle detection" in *International Conference on Multimedia and Expo*, 2017, pp. 1135–1140.

[48] S. Lyu *et al.*, "UA-DETRAC 2017: Report of AVSS2017 & IWT4S Challenge on Advanced Traffic Monitoring" in *2017 14th IEEE International Conference on Advanced Video and Signal Based Surveillance (AVSS)*, 2017, pp. 1–7.